\documentclass{edm_article}
\pdfoutput=1

\usepackage{csquotes}
\usepackage{todonotes}
\setuptodonotes{inline}
\usepackage{microtype}
\usepackage{url}
\usepackage{cite}
\usepackage[numbers]{natbib}
\usepackage[colorlinks,allcolors=blue]{hyperref}
\usepackage[capitalise,noabbrev]{cleveref}
\creflabelformat{equation}{#2\textup{#1}#3}

\newtheorem{definition}{Definition}

\begin{document}

\title{Towards Equalised Odds as Fairness Metric in Academic Performance Prediction}
\numberofauthors{2}
\author{
\alignauthor
Jannik Dunkelau\\
  \affaddr{Heinrich-Heine-Universit\"at D\"usseldorf}\\
  \affaddr{D-40225 D\"usseldorf, Germany}\\
  \email{jannik.dunkelau@hhu.de}
\alignauthor
Manh Khoi Duong\\
  \affaddr{Heinrich-Heine-Universit\"at D\"usseldorf}\\
  \affaddr{D-40225 D\"usseldorf, Germany}\\
  \email{manh.khoi.duong@hhu.de}
}
\date{27th July 2022}

\maketitle

\begin{abstract}
  The literature for fairness-aware machine learning knows a plethora
  of different fairness notions.
  It is however well-known, that it is impossible to satisfy all of them,
  as certain notions contradict each other.
  In this paper, we take a closer look at academic performance prediction (APP)
  systems and try to distil which fairness notions suit this task most.
  For this,
  we scan recent literature proposing guidelines as to which fairness notion
  to use and apply these guidelines onto APP.
  Our findings suggest equalised odds as most suitable notion for APP,
  based on APP's WYSIWYG worldview as well as potential long-term
  improvements for the population.
\end{abstract}


\section{Introduction}
Socially responsible and fairness-aware machine learning (FairML)
is becoming
increasingly more important to our society and
aggregated a large body of research regarding
how to ensure fairness and non-discrimination by
artificially intelligent system~%
\cite{friedler2019comparative,dunkelau2019fairness,zliobaite2017measuring,pessach2020algorithmic,mehrabi2021survey}.
As a consequence,
the notion of FairML found its way into the research
of educational recommender systems as well wherever
a social impact onto the student body is at stake~%
\cite{loukina2019many,kizilcec2020algorithmic,keller2022how}.
A major part in this plays
academic performance prediction (APP).
Hereby, an APP system takes data of a student as input,
predicts how the student will perform in the future,
and may hence induce an action based on this prediction which may
itself affect the student~\cite{alyahyan2020predicting}.
Such predictions are usually employed
as early-warning system to determine students at risk,
intervene by supplying necessary help and resources,
and increase graduation rates as a consequence~%
\cite{daniel2014big,alyahyan2020predicting,attaran2018opportunities,keller2022how}.
Although other utilisation of APP is possible,
e.g.\ guiding university admissions,
we will focus on the use case of targeted interventions.
Given the need for socially responsible APP systems~%
\cite{keller2022how,kizilcec2020algorithmic},
the question arises as to which notion of fairness is suitable for APP\@.

In the following,
we review literature regarding selection of fairness notions,
derive a reduced guideline to decide between
two popular, parity-based fairness notions,
demographic parity and equalised odds, and
apply our findings onto APP\@.
Our results and main contributions are the relationship of
APPs to equalised odds and the WYSIWYG worldview
which is backed by literature.
Motivated by own work regarding the conceptualisation of responsible APP,
we hope to narrow down the research focus for APP fairness notions,
provide a base-notion for new and established APP researchers alike,
and to contribute to public discourse on this matter.

\section{Notation}


In the following,
let \(\mathcal{D} = \{(x_i, y_i, z_i)\}_{i=1}^n \subset
\mathbf{X}\times\mathbf{Y}\times\mathbf{Z}\),
denote the training data
of individuals
where
\(\mathbf{X}\subset\mathbb{R}^d\) is the \(d\)-dimensional set of input (feature) vectors
characterising each individual,
\(\mathbf{Y}\) denotes the set of measured true labels over the individuals,
and \(\mathbf{Z}\) is the set of protected attributes corresponding to each
individual.
Given a classifier \(h\), we denote the set of its predictions over
\(\mathbf{X}\) as \(\mathbf{\hat{Y}}\).
Without loss of generality, we assume
\(y\in\mathbf{Y}\) to be binary in \(\{0,1\}\).
We say an individual \((x,y,z) \in \mathcal{D}\)
receives the \emph{favourable outcome} if the prediction \(\hat{y}=h(x)=1\).
Otherwise, we say the individual receives the \emph{unfavourable outcome}.
We say the individual belongs to the demographic group \(z\).
Further,
let \(X, Y, Z, \hat{Y}\) denote random variables
describing the events that, for an individual from the training data,
their features, ground truth, protected attributes, and classifier prediction
take a specific value, respectively.
Thus, \(P(\hat{Y}=1\mid Y=1)\) denotes the probability
that individuals with a positive ground truth are receiving the
favourable outcome.

\section{Parity-based Fairness Notions}%
\label{sec:notions}

Parity-based fairness notions are defined over the values of a classifier's
confusion matrix~\cite{corbettdavies2018measure}.
They assume fairness once a set of predictive rates is equal for each
demographic group,
for instance the positive prediction rate,
true positive prediction rate, or false positive prediction rate,
as we will see below.
For this work,
we focus on two such notions which are currently prevalent
in literature: demographic parity and equalised odds.
We selected these notions as they seem to have higher citation counts as others%
~\cite{verma2018fairness}
and are accounted for by related literature as well%
~\cite{corbettdavies2018measure,kizilcec2020algorithmic,pessach2020algorithmic}.

Demographic parity assumes that the distribution of the favourable outcome
should be equal throughout all demographic groups.
It is formally defined in \cref{def:demographic-parity}:

\begin{definition}[Demographic Parity]\label{def:demographic-parity}
  We say that a classifier satisfies \emph{demographic parity}
  if the positive prediction rate is equal for all demographic groups,
  i.e.
  \begin{equation}
    P(\hat{Y}=1\mid Z=z) = P(\hat{Y}=1)
    \,\text{.}
  \end{equation}
\end{definition}

While demographic parity is the most popular fairness metric in literature,
it also exhibits various short comings.
For instance, randomizing predictions for one demographic group while having
proper predictions for another can already satisfy the notion~%
\cite{dwork2012fairness}.
It is however independent from any possible bias in the collection of
the ground truth values \(\mathbf{Y}\) which could have been assembled
in a discriminatory way~\cite{barocas2017big}
as the notion does not rely on \(\mathbf{Y}\) at all.

As an alternative, \citet{hardt2016equality}
proposed the notion of equalised odds as given in \cref{def:equalised-odds},
which assumes fairness if \(\hat{Y}\perp Z\mid Y\).
As equalised odds is defined over true and false positive rates of a classifier,
it is always satisfied if \(\mathbf{\hat{Y}}=\mathbf{Y}\)
which is not guaranteed for demographic parity.

\begin{definition}[Equalised Odds]\label{def:equalised-odds}
  We say that a classifier satisfies \emph{equalised odds}
  if it has equal true positive rates and false positive rates
  for all demographic groups,
  i.e.
  \begin{gather}
    P(\hat{Y}=1\mid {Y}=1, Z=z) = P(\hat{Y}=1\mid {Y}=1)\label{eq:tpr} \\
    P(\hat{Y}=1\mid {Y}=0, Z=z) = P(\hat{Y}=1\mid {Y}=0)\label{eq:fpr}
  \end{gather}
\end{definition}

\section{Worldviews}

Recent literature promotes accounting for the worldview that underlies the data%
~\cite{friedler2016possibility,yeom2021wordviews,kizilcec2020algorithmic,makhlouf2021machine}.
Worldviews were introduced by
\citet{friedler2016possibility}.
To define them,
we must firstly differentiate between the
\emph{observable space} \(\mathcal{Y}\) and the
\emph{construct space} \(\mathcal{Y'}\).
The observable space \(\mathcal{Y}\) represents the room of available
observations and measurements.
The training data \(\mathcal{D}\) can only be collected from the observable
space.
On the other hand,
the construct space \(\mathcal{Y'}\)
represents the theoretical space of the \enquote{true} data
that is relevant to the task but not measurable.
For instance,
assume the task to predict whether a student will graduate within the
standard duration of study.
We can collect historical information of graduates to model the target variable
\(Y\) and select characteristics such as grades within the first semester
or number of passed courses per semester as features.
These are part of the observable space that is available to us.
The related construct space would contain information about
how well passed courses were understood,
how motivated the students will remain long-term,
or how much time they will be able to invest into their studies in later
semesters. This information is not accessible to us but can only be observed
via assumed proxies.
Further more,
the construct space is free from discrimination in a sense that it would
not contain the grades a student received but rather the grade a student should
have received if no discrimination took place.

Worldviews model the expected differences between demographic groups
in \(\mathcal{Y'}\) and hence explain the presence of measured differences in
\(\mathcal{Y}\)~\cite{yeom2021wordviews}.
Two prominent worldviews are \emph{We're all equal} and
\emph{What you see is what you get},
for which we borrow
\cref{def:wae,def:wysiwyg}
of \citet{yeom2021wordviews}.
Both where originally formulated by \citet{friedler2016possibility}
and seem to represent two polar ends in the fairness literature.

\begin{definition}[WAE]\label{def:wae}
  \emph{We're all equal (WAE)} represents a world view which assumes that each demographic group
  is identical to each other with respect to the target variable
  in the construct space, i.e. \(\mathcal{Y'}\perp Z\).
\end{definition}

\begin{definition}[WYSIWYG]\label{def:wysiwyg}
  \emph{What you see is what you get (WYSIWYG)} is a worldview which
  assumes that differences in \(\mathcal{Y}\)
  are explained by differences in \(\mathcal{Y'}\)
  and hence that the observable space is an accurate reflection of the
  construct, i.e. \(\mathcal{Y}=\mathcal{Y'}\).
\end{definition}

As we consider WAE and WYSIWYG in contexts in which we do observe discrimination
in the observable space \(\mathcal{Y}\) and thus assume
\(\mathcal{Y}\not\perp Z\), both world views contradict each other
in context of this work.

\section{Fairness Selection Guidelines}%
\label{sec:notion-selection}

While literature produced a great number of fairness notions to choose from,
we know different fairness notions to be mutually
exclusive to one another,
making it impossible to satisfy all notions simultaneously~%
\cite{friedler2016possibility,pleiss2017fairness,kleinberg2016inherent,chouldechova2017fair}.
Specifically,
the notions from \cref{sec:notions} above are mutually exclusive
in non-trivial cases.
Hence,
it is valuable to know which fairness metric suits the prediction task most.

\citet{makhlouf2021machine}
collected a decision diagram guiding the fairness notion selection process.
This diagram leads to the selection of
demographic parity if standards exist which regulate the ratio of admission
rates for the favourable outcome or we do not have a reliable ground truth
or can assume historical bias or measurement bias in the data.
Further,
when we have a reliable ground truth or assume no historical or measurement
biases in our data,
the authors advocate for equalised odds if the emphasis is on
the classification recall.
\citeauthor{makhlouf2021machine} further advance the idea that
the selection of fairness notion must be based on the explicit choice of
an underlying worldview.
The worldview itself is however not (explicitly) part of their guiding diagram.
If we however focus on the distinction between reliability of \(\mathbf{Y}\)
(and existence or absence of biases),
we can infer that a reliable \(\mathbf{Y}\) relates to
\(\mathcal{Y}\approx\mathcal{Y'}\) and thus WYSIWYG, and an unreliable
\(\mathbf{Y}\) relates hence to WAE\@.

\citet{friedler2016possibility}
show in their initial conception of worldviews
that individual fairness can be guaranteed under WYSIWYG while it can
cause discrimination in a WAE setting.
On the flip side,
demographic parity is not applicable in a WYSIWYG setting
while it can guarantee non-discrimination for WAE\@.
\citet{yeom2021wordviews} investigated the theoretical impact
the selection of a fairness notion has on the disparity between groups.
In their work,
they prove that any model that satisfies demographic parity on
\(\mathbf{\hat{Y}}\) does not amplify existing disparity in \(\mathcal{Y'}\).
However, only WAE lends itself to demographic parity,
as the classification performance with respect to \(\mathcal{Y'}\)
in WYSIWYG will always be suboptimal.
A model satisfying equalised odds will not amplify any disparity in WYSIWYG
but can amplify disparities if WAE holds.

Unifying the guidelines and insights from above,
demographic parity should be employed when WAE holds.
That is,
we desire an equal distribution of the favourable outcome throughout the groups
as we accredit any measurable differences in our training data
to prior discrimination (historical or elsewise).
Equalised odds should be favoured
if WYSIWYG holds.
That is, we expect differences between groups to be explainable by
differences in the construct space \(\mathcal{Y'}\).
Some literature also promotes demographic parity
when the \emph{playing field is even} for the groups~%
\cite{kizilcec2020algorithmic,deho2022how}
or the classifier is employed for one group independently~\cite{loukina2019many}
while promoting equalised odds otherwise~%
\cite{kizilcec2020algorithmic,loukina2019many}.

\section{Towards an APP Fairness Notion}

In this section,
we will discuss the worldview that generally seems to tie to APP systems,
derive equalised odds as the appropriate fairness notion,
then take a closer look at the benefits and drawbacks equalised odds exhibits.
We conclude with a brief overview of selected notions which
we did not consider in depth.

\subsection{The APP Worldview}

To evaluate which worldview gives itself to APP systems,
we investigate below which input features
and target variables
promote which worldview
to conclude the related fairness notion.
For this,
we lean on the work of
\citet{alyahyan2020predicting},
who report the mostly used influential features for APP
to be
prior academic achievement and student demographics,
accounting for 70 \% of their surveyed articles.

Prior academic achievement is mostly concerned with grade-related features
which are aggregated during university~%
\cite{alyahyan2020predicting}:
specific course grades, grade point average (GPA),
cumulative GPA, exam results, or individual assessment grades;
but also pre-university features such as high school background
or study admission test results.

Taking grade-related features into account to predict on graduation level,
it feels intuitive that we are in a WYSIWYG environment.
Not because the grading of students can be assumed to be unbiased
(which it cannot, cf.~\cite{malouff2014preventing,malouff2016bias}),
but because once the grades are set,
different impact onto the graduation level prediction can be solely explained
by different grade distributions.
For instance,
assume the task to predict qualification for a subsequent master's programme.
The qualification is decided by achieving a certain GPA at graduation.
As the grade-based input features are already set,
final GPA is rendered to a consequence
and disparities can be explained by differences in cumulative
grades.

The same argument can be made for
using student demographics as features.
Hereby, student demographics refer to protected attributes such as gender,
race, religion, or socioeconomic status~\cite{alyahyan2020predicting}.
In a discriminatory system which grades minority groups worse than
privileged groups,
the protected attribute effects achieving lower grades,
again rendering final GPA as a consequence.
Hence, WYSIWYG holds,
explaining outcome disparities due to membership in certain demographic groups.
Despite this very discriminatory interpretation,
WAE is not an applicable worldview in that scenario:
If we assume merit to be equally distributed throughout all demographic groups,
it generally will not hold that unevenly distributed cumulative GPAs
should result in equally distributed final GPAs.

The above observations indicate that
APP assumes WYSIWYG\@.
This can further be supported by the following two argumentations.
Firstly, due to unequal distribution of resources among demographic groups,
educational disparities are to be expected~\cite{american1999standards}.
Secondly,
there is a difference between ideal and non-ideal fairness-perspectives~%
\cite{fazelpour2020algorithmic}.
The fairness ideal would imply that grade-level outcomes are equally distributed
throughout groups.
Our world is however non-ideal and the fairness ideal is the target state
we aim to achieve.
For this, we measure the deviation of our systems from the fairness ideal
in FairML and attempt to minimise it~\cite{fazelpour2020algorithmic}.

As WYSIWYG for APP
seems to find support in literature,
consequentially APP pairs with the fairness notion of equalised odds.
This aligns with (and generalises) the statement of
\citet{kizilcec2020algorithmic} that equalised odds is
\enquote{most appropriate in applications like student dropout prediction}.
Having singled out equalised odds as fairness notion,
we will inspect its suitability further and discard demographic parity
in the remainder of this paper.

\subsection{A Closer Look at Equalised Odds}

While we identified equalised odds as a fairness notion which
pairs well with APP, there are further concerns in literature
regarding the fairness notion of a FairML system which remain to
be discussed. 
\citet{fazelpour2020algorithmic}
note that the approach to FairML should consider
situated and system wide as well as dynamic impacts of APP intervention
while \citet{deho2022how}
promote to focus on equity and need rather than equality.

\paragraph{Favourable outcome revisited}
In classical FairML,
we assume \(\hat{y}=1\) to denote a favourable outcome,
such as an approved credit loan or being hired at a new job.
Intuitively,
the favourable outcome in APP for a student is to be predicted
as a successful student.
However,
the real classification task behind APP is rather
to predict the need of intervention to help the student achieve a higher
performance.
The emphasis from a stakeholder's perspective lies on the need of action.
Thus we can reframe the favourable outcome in APP as
dependent on \({Y}\).
For at-risk students with \(y=1\)
the favourable outcome is indeed \(\hat{y}=1\) so they receive the intervention.
For students who will graduate without further intervention and thus \(y=0\)
the favourable outcome would be to not get flagged as at-risk,
i.e.~\(\hat{y}=0\).
Thus, for APP, the favourable outcome would be a perfect predictor with
\(\hat{y}=y\).
Such a predictor would always satisfy equalised odds~\cite{hardt2016equality}.
This differs from classical FairML as
the students did not apply for the interventions,
contrasting loan or job applications
where we assume an approved application to be favoured by the individual.

\paragraph{Long-term impacts}
\citet{liu2018delayed} show that both,
demographic parity as well as equal true positive rates
(only \cref{eq:tpr} from equalised odds satisfied),
are able to cause improvement, stagnation,
or even decline in the long-term well-being of disadvantaged groups,
depending on the settings.
While not considering the stricter notion of equalised odds,
their results still suggest that further inspection of
respectively underlying distributions of \(Y\) needs to be accounted for.
Contrasting this with the results of \citet{yeom2021wordviews} however,
that equalised odds will not amplify discrimination when WYSIWYG holds,
gives at least some kind of (theoretical) reassurance of the selection of
equalised odds as fairness notion.
Further,
due to the intervening nature of APP
as well as the favourable outcome being dependent on \(Y\),
we can illustrate at least a partial improvement over time.
As stated above,
educational disparities are to be expected due to resources being
unequally distributed and our world being non-ideal~%
\cite{american1999standards,fazelpour2020algorithmic}.
Hence,
we can assume a proportionally higher rate of \(y=1\) in minority groups.
For an APP system satisfying equalised odds,
this would result in a higher proportion of minority students receiving
the intervention.
Assuming the intervention increases graduation rate and/or graduation quality,
it should increase the availability of resources for these groups long-term.
Thus, the divergence from the fairness-ideal should be reduced.
This however only narrows the gap but will be unable to close it,
as for instance biases in grading may not be cured in this process.

\paragraph{From Equality to Equity}

Instead of promoting equal treatment as measure of fairness,
\citet{deho2022how} propose to focus on equity and needed treatment instead.
However, it is unclear from their paper whether they regard equalised odds
to be a measure of equity,
whereby
\citet{jiang2021towards} apply data rebalancing techniques to boost equity
for an APP system
in terms of true positive and true negative rates,
hence they use equalised odds as measure for equity.
This makes sense for APP, as the intervening nature inherently
attempts to target students at risk.
However, \citet{naggita2022equity} show that
a system satisfying equalised odds can still promote inequity.
This is conditioned over the accessibility of the system towards the
demographic groups.
\emph{Accessibility} is hereby defined over the notion of obstacles
which obstruct an individual to exhibit their true feature vector towards
the prediction system.
Such obstructions could be due to biased grading processes
which APP alone is unable to solve.

\paragraph{Limitations}
\citet{corbettdavies2018measure}
have shown that equalised odds, as well as all parity based notions,
is subject to the problem of infra-marginality
as a unified classification threshold is not sensible
if the underlying risk distributions are unequal for two demographic groups.
In such cases, the error scores will differ and parity cannot be achieved.
Furthermore,
equalised odds is usually only satisfiable when different classification
thresholds for the demographic groups are employed in the first place~%
\cite{gardner2019evaluating,kizilcec2020algorithmic}.
In such cases, the use of the protected attribute is needed at classification
time,
which might not everywhere be legally feasible.
However,
\citet{yu2021should} argue that APP systems such as dropout detection
should include protected attributes,
albeit the authors only report a limited benefit in terms of
fairness and performance.

\paragraph{Students' Perceived Fairness}
First work analyses the implications and perceptions of
fairness in APP systems~\cite{marcinkowski2020implications,smith2017students},
however
a more thorough investigation regarding equalised odds needs yet to be
conducted.
While \citet{smith2017students} report student's focus on
relational and stake fairness, which equalised odds could cater to,
\citet{marcinkowski2020implications}
report focus on distributional and procedural fairness dimensions.
Although equalised odds fits procedural fairness,
it fails to do so for distributional fairness which would rather be satisfied
by demographic parity instead.
This could be overcome by a weighted trade-off between both notions
as suggested by \citet{kizilcec2020algorithmic}.
However, it is unclear whether the benefits of equalised odds remain
unaffected in this case
or whether the student body is willing of such a compromise.

\subsection{Undiscussed Notions}

We only described two fairness notions in \cref{sec:notions},
but current literature provides a plethora of further notions~%
\cite{verma2018fairness,mehrabi2021survey,dunkelau2019fairness,makhlouf2021machine}
While it is not possible for us to talk about all of them,
we will highlight selected notions and outline their relevance for APP
or why we discarded them in our work.

Next to demographic parity and equalised odds,
calibration~\cite{pleiss2017fairness}
and predictive parity are also popular notions in
literature.
However,
\citet{yeom2021wordviews} showed that neither WAE nor WYSIWYG motivate
either notion.

Work that considers worldviews usually promotes individual fairness~%
\cite{dwork2012fairness}
as suitable for a WYSIWYG setting~%
\cite{friedler2016possibility,kizilcec2020algorithmic}.
Individual fairness is strictly not parity based,
but we intended to review parity based notions specifically.
However, as both, equalised odds and individual fairness,
are promoted for WYSIWYG settings, an investigation
of their relationship should be followed up in future work.

\citet{gardner2019evaluating} introduced ABROCA scores as measure for fairness,
which rely on different ROC curves of the demographic groups.
While equalised odds is satisfied at intersections of ROC curves,
slicing analysis with ABROCA allows for a more nuanced evaluation of the
overall fairness trends for different classification thresholds.
Specifically,
if one does not require equality for the demographic groups in
\cref{eq:fpr,eq:tpr} but only requires an absolute difference of at most
\(\epsilon\), ABROCA might allow for easier selection of classification
thresholds.
Whether guarantees regarding disparity amplification under WYSIWYG stay
true is left for future work.

\citet{yeom2021wordviews}
define the notion of an \(\alpha\)-hybrid worldview
which assumes that discrimination in \(\mathcal{Y}\)
is partially explained in \(\mathcal{Y'}\) to a ration of \(\alpha\in[0,1]\)
and thus positions itself between WAE and WYSIWYG\@.
While the authors present the \(\alpha\)-disparity test as a fairness measure,
the value of \(\alpha\) needs to be approximated by social research as well as
public discourse.


\section{Conclusions}
In this work
we reviewed recent literature in search of finding a suitable fairness notion
to employ in responsible APP systems.
The consensus of our search favours equalised odds over demographic parity,
calibration, or predictive parity.
After highlighting APPs relation to WYSIWYG,
we further found support of equalised odds in terms of
reframing the favourable outcome, inspecting possible long-term impacts
and partly relating to equity notions.
While equalised odds still shows limitations in its applicability,
we emphasise the need of further analysis regarding equalised odds in APP
contexts specifically:
in terms of equity, relation to individual fairness, and perceived fairness.

\section{Acknowledgments}
This work was supported by the Federal Ministry of Education and Research
(BMBF) under Grand No.~16DHB4020.

%
\bibliographystyle{abbrvnat}
\bibliography{sigproc}  

\begin{thebibliography}{34}
\providecommand{\natexlab}[1]{#1}
\providecommand{\url}[1]{\texttt{#1}}
\expandafter\ifx\csname urlstyle\endcsname\relax
  \providecommand{\doi}[1]{doi: #1}\else
  \providecommand{\doi}{doi: \begingroup \urlstyle{rm}\Url}\fi

\bibitem[AERA(1999)]{american1999standards}
AERA.
\newblock \emph{Standards for educational and psychological testing}.
\newblock American Educational Research Association, 1999.

\bibitem[Alyahyan and Dü{\c{s}}tegör(2020)]{alyahyan2020predicting}
E.~Alyahyan and D.~Dü{\c{s}}tegör.
\newblock Predicting academic success in higher education: literature review
  and best practices.
\newblock \emph{International Journal of Educational Technology in Higher
  Education}, 17\penalty0 (1), feb 2020.
\newblock \doi{10.1186/s41239-020-0177-7}.

\bibitem[Attaran et~al.(2018)Attaran, Stark, and
  Stotler]{attaran2018opportunities}
M.~Attaran, J.~Stark, and D.~Stotler.
\newblock Opportunities and challenges for big data analytics in {US} higher
  education.
\newblock \emph{Industry and Higher Education}, 32\penalty0 (3):\penalty0
  169--182, apr 2018.
\newblock \doi{10.1177/0950422218770937}.

\bibitem[Barocas et~al.(2017)Barocas, Bradley, Honavar, and
  Provost]{barocas2017big}
S.~Barocas, E.~Bradley, V.~Honavar, and F.~Provost.
\newblock Big data, data science, and civil rights.
\newblock \emph{arXiv preprint arXiv:1706.03102}, 2017.

\bibitem[Chouldechova(2017)]{chouldechova2017fair}
A.~Chouldechova.
\newblock Fair prediction with disparate impact: A study of bias in recidivism
  prediction instruments.
\newblock \emph{Big data}, 5\penalty0 (2):\penalty0 153--163, 2017.

\bibitem[Corbett-Davies and Goel(2018)]{corbettdavies2018measure}
S.~Corbett-Davies and S.~Goel.
\newblock The measure and mismeasure of fairness: A critical review of fair
  machine learning.
\newblock July 2018.

\bibitem[Daniel(2014)]{daniel2014big}
B.~Daniel.
\newblock Big data and analytics in higher education: Opportunities and
  challenges.
\newblock \emph{British Journal of Educational Technology}, 46\penalty0
  (5):\penalty0 904--920, dec 2014.
\newblock \doi{10.1111/bjet.12230}.

\bibitem[Deho et~al.(2022)Deho, Zhan, Li, Liu, Liu, and Le]{deho2022how}
O.~B. Deho, C.~Zhan, J.~Li, J.~Liu, L.~Liu, and T.~D. Le.
\newblock How do the existing fairness metrics and unfairness mitigation
  algorithms contribute to ethical learning analytics?
\newblock \emph{British Journal of Educational Technology}, apr 2022.
\newblock \doi{10.1111/bjet.13217}.

\bibitem[Dunkelau and Leuschel(2019)]{dunkelau2019fairness}
J.~Dunkelau and M.~Leuschel.
\newblock Fairness-aware machine learning: An extensive overview.
\newblock Working paper, available at
  \url{https://www3.hhu.de/stups/downloads/pdf/fairness-survey.pdf}, Oct. 2019.

\bibitem[Dwork et~al.(2012)Dwork, Hardt, Pitassi, Reingold, and
  Zemel]{dwork2012fairness}
C.~Dwork, M.~Hardt, T.~Pitassi, O.~Reingold, and R.~Zemel.
\newblock Fairness through awareness.
\newblock In \emph{Proceedings of the 3rd innovations in theoretical computer
  science conference}, pages 214--226. ACM, 2012.

\bibitem[Fazelpour and Lipton(2020)]{fazelpour2020algorithmic}
S.~Fazelpour and Z.~C. Lipton.
\newblock Algorithmic fairness from a non-ideal perspective.
\newblock In \emph{Proceedings of the {AAAI}/{ACM} Conference on {AI}, Ethics,
  and Society}. {ACM}, feb 2020.
\newblock \doi{10.1145/3375627.3375828}.

\bibitem[Friedler et~al.(2016)Friedler, Scheidegger, and
  Venkatasubramanian]{friedler2016possibility}
S.~A. Friedler, C.~Scheidegger, and S.~Venkatasubramanian.
\newblock On the (im)possibility of fairness.
\newblock \emph{arXiv preprint arXiv:1609.07236}, 2016.

\bibitem[Friedler et~al.(2019)Friedler, Scheidegger, Venkatasubramanian,
  Choudhary, Hamilton, and Roth]{friedler2019comparative}
S.~A. Friedler, C.~Scheidegger, S.~Venkatasubramanian, S.~Choudhary, E.~P.
  Hamilton, and D.~Roth.
\newblock A comparative study of fairness-enhancing interventions in machine
  learning.
\newblock In \emph{Proceedings of the Conference on Fairness, Accountability,
  and Transparency}. {ACM}, jan 2019.
\newblock \doi{10.1145/3287560.3287589}.

\bibitem[Gardner et~al.(2019)Gardner, Brooks, and Baker]{gardner2019evaluating}
J.~Gardner, C.~Brooks, and R.~Baker.
\newblock Evaluating the fairness of predictive student models through slicing
  analysis.
\newblock In \emph{Proceedings of the 9th International Conference on Learning
  Analytics and Knowledge}. {ACM}, mar 2019.
\newblock \doi{10.1145/3303772.3303791}.

\bibitem[Hardt et~al.(2016)Hardt, Price, Srebro, et~al.]{hardt2016equality}
M.~Hardt, E.~Price, N.~Srebro, et~al.
\newblock Equality of opportunity in supervised learning.
\newblock In \emph{Advances in neural information processing systems}, pages
  3315--3323, 2016.

\bibitem[Jiang and Pardos(2021)]{jiang2021towards}
W.~Jiang and Z.~A. Pardos.
\newblock Towards equity and algorithmic fairness in student grade prediction.
\newblock In \emph{Proceedings of the 2021 {AAAI}/{ACM} Conference on {AI},
  Ethics, and Society}. {ACM}, jul 2021.
\newblock \doi{10.1145/3461702.3462623}.

\bibitem[Keller et~al.(2022)Keller, Lünich, and Marcinkowski]{keller2022how}
B.~Keller, M.~Lünich, and F.~Marcinkowski.
\newblock How is socially responsible academic performance prediction possible?
\newblock In \emph{Strategy, Policy, Practice, and Governance for {AI} in
  Higher Education Institutions}, pages 126--155. {IGI} Global, may 2022.
\newblock \doi{10.4018/978-1-7998-9247-2.ch006}.

\bibitem[Kizilcec and Lee(2020)]{kizilcec2020algorithmic}
R.~F. Kizilcec and H.~Lee.
\newblock Algorithmic fairness in education.
\newblock arXiv, 2020.
\newblock \doi{10.48550/ARXIV.2007.05443}.

\bibitem[Kleinberg et~al.(2017)Kleinberg, Mullainathan, and
  Raghavan]{kleinberg2016inherent}
J.~Kleinberg, S.~Mullainathan, and M.~Raghavan.
\newblock Inherent trade-offs in the fair determination of risk scores.
\newblock In C.~H. Papadimitriou, editor, \emph{8th Innovations in Theoretical
  Computer Science Conference (ITCS 2017)}, volume~67 of \emph{Leibniz
  International Proceedings in Informatics (LIPIcs)}, pages 43:1--43:23,
  Dagstuhl, Germany, 2017. Schloss Dagstuhl--Leibniz-Zentrum fuer Informatik.

\bibitem[Liu et~al.(2018)Liu, Dean, Rolf, Simchowitz, and
  Hardt]{liu2018delayed}
L.~T. Liu, S.~Dean, E.~Rolf, M.~Simchowitz, and M.~Hardt.
\newblock Delayed impact of fair machine learning.
\newblock In J.~Dy and A.~Krause, editors, \emph{Proceedings of the 35th
  International Conference on Machine Learning}, volume~80 of \emph{Proceedings
  of Machine Learning Research}, pages 3150--3158. PMLR, 10--15 Jul 2018.
\newblock URL \url{https://proceedings.mlr.press/v80/liu18c.html}.

\bibitem[Loukina et~al.(2019)Loukina, Madnani, and Zechner]{loukina2019many}
A.~Loukina, N.~Madnani, and K.~Zechner.
\newblock The many dimensions of algorithmic fairness in educational
  applications.
\newblock In \emph{Proceedings of the Fourteenth Workshop on Innovative Use of
  {NLP} for Building Educational Applications}, pages 1--10. Association for
  Computational Linguistics, Aug. 2019.
\newblock \doi{10.18653/v1/w19-4401}.

\bibitem[Makhlouf et~al.(2021)Makhlouf, Zhioua, and
  Palamidessi]{makhlouf2021machine}
K.~Makhlouf, S.~Zhioua, and C.~Palamidessi.
\newblock Machine learning fairness notions: Bridging the gap with real-world
  applications.
\newblock \emph{Information Processing \& Management}, 58\penalty0
  (5):\penalty0 102642, sep 2021.
\newblock \doi{10.1016/j.ipm.2021.102642}.

\bibitem[Malouff and Thorsteinsson(2016)]{malouff2016bias}
J.~M. Malouff and E.~B. Thorsteinsson.
\newblock Bias in grading: A meta-analysis of experimental research findings.
\newblock \emph{Australian Journal of Education}, 60\penalty0 (3):\penalty0
  245--256, sep 2016.
\newblock \doi{10.1177/0004944116664618}.

\bibitem[Malouff et~al.(2014)Malouff, Stein, Bothma, Coulter, and
  Emmerton]{malouff2014preventing}
J.~M. Malouff, S.~J. Stein, L.~N. Bothma, K.~Coulter, and A.~J. Emmerton.
\newblock Preventing halo bias in grading the work of university students.
\newblock \emph{Cogent Psychology}, 1\penalty0 (1):\penalty0 988937, dec 2014.
\newblock \doi{10.1080/23311908.2014.988937}.

\bibitem[Marcinkowski et~al.(2020)Marcinkowski, Kieslich, Starke, and
  Lünich]{marcinkowski2020implications}
F.~Marcinkowski, K.~Kieslich, C.~Starke, and M.~Lünich.
\newblock Implications of {AI} (un-)fairness in higher education admissions.
\newblock In \emph{Proceedings of the 2020 Conference on Fairness,
  Accountability, and Transparency}. {ACM}, jan 2020.
\newblock \doi{10.1145/3351095.3372867}.

\bibitem[Mehrabi et~al.(2021)Mehrabi, Morstatter, Saxena, Lerman, and
  Galstyan]{mehrabi2021survey}
N.~Mehrabi, F.~Morstatter, N.~Saxena, K.~Lerman, and A.~Galstyan.
\newblock A survey on bias and fairness in machine learning.
\newblock \emph{{ACM} Computing Surveys}, 54\penalty0 (6):\penalty0 1--35, jul
  2021.
\newblock \doi{10.1145/3457607}.

\bibitem[Naggita and Aguma(2022)]{naggita2022equity}
K.~Naggita and J.~C. Aguma.
\newblock The equity framework: Fairness beyond equalized predictive outcomes.
\newblock arXiv, 2022.
\newblock \doi{10.48550/ARXIV.2205.01072}.

\bibitem[Pessach and Shmueli(2020)]{pessach2020algorithmic}
D.~Pessach and E.~Shmueli.
\newblock Algorithmic fairness.
\newblock volume abs/2001.09784, 2020.

\bibitem[Pleiss et~al.(2017)Pleiss, Raghavan, Wu, Kleinberg, and
  Weinberger]{pleiss2017fairness}
G.~Pleiss, M.~Raghavan, F.~Wu, J.~Kleinberg, and K.~Q. Weinberger.
\newblock On fairness and calibration.
\newblock In \emph{Advances in Neural Information Processing Systems}, pages
  5680--5689, 2017.

\bibitem[Smith et~al.(2017)Smith, Todd, and Laing]{smith2017students}
L.~M. Smith, L.~Todd, and K.~Laing.
\newblock Students' views on fairness in education: the importance of
  relational justice and stakes fairness.
\newblock \emph{Research Papers in Education}, 33\penalty0 (3):\penalty0
  336--353, mar 2017.
\newblock \doi{10.1080/02671522.2017.1302500}.

\bibitem[Verma and Rubin(2018)]{verma2018fairness}
S.~Verma and J.~Rubin.
\newblock Fairness definitions explained.
\newblock In \emph{FairWare'18: IEEE/ACM International Workshop on Software
  Fairness}. ACM, May 2018.

\bibitem[Yeom and Tschantz(2021)]{yeom2021wordviews}
S.~Yeom and M.~C. Tschantz.
\newblock Avoiding disparity amplification under different worldviews.
\newblock In \emph{Conference on Fairness, Accountability, and Transparency},
  FAccT '21, page 273–283, New York, NY, USA, 2021. ACM.
\newblock \doi{10.1145/3442188.3445892}.

\bibitem[Yu et~al.(2021)Yu, Lee, and Kizilcec]{yu2021should}
R.~Yu, H.~Lee, and R.~F. Kizilcec.
\newblock Should college dropout prediction models include protected
  attributes?
\newblock In \emph{Proceedings of the Eighth {ACM} Conference on Learning @
  Scale}. {ACM}, jun 2021.
\newblock \doi{10.1145/3430895.3460139}.

\bibitem[{\v{Z}}liobait{\.{e}}(2017)]{zliobaite2017measuring}
I.~{\v{Z}}liobait{\.{e}}.
\newblock Measuring discrimination in algorithmic decision making.
\newblock \emph{Data Mining and Knowledge Discovery}, 31\penalty0 (4):\penalty0
  1060--1089, July 2017.

\end{thebibliography}
%
\balancecolumns
\end{document}